\pdfoutput=1

\documentclass[11pt]{article}

\usepackage{emnlp2021}

\usepackage{times}
\usepackage{latexsym}
\usepackage{booktabs}
\usepackage{graphicx, verbatim}
\usepackage[disable]{todonotes}
\usepackage{amsmath, amssymb}
\usepackage{balance}
\usepackage[noend,ruled]{algorithm2e}
\DeclareMathOperator*{\argmax}{arg\,max}

\usepackage[T1]{fontenc}

\usepackage[utf8]{inputenc}

\usepackage{microtype}

\title{Fixing exposure bias with imitation learning needs powerful oracles}

\author{Luca Hormann$^\ast$ \and Artem Sokolov$^{\bullet,\ast}$\\
  Heidelberg University$^\ast$, Google Research$^\bullet$\\
  \texttt{\{luca.hormann@stud, sokolov@cl\}.uni-heidelberg.de}}

\begin{document}
\maketitle
\begin{abstract}
We apply imitation learning (IL) to tackle the NMT exposure bias problem with error-correcting oracles, and evaluate an SMT lattice-based oracle which, despite its excellent performance in an unconstrained oracle translation task, turned out to be too pruned and idiosyncratic to serve as the oracle for IL.
\end{abstract}


\todo{change the name of the Overleaf project to the final title}
\section{Introduction}

Catastrophic failures of neural machine translation (NMT) systems -- hallucinations, repeated or nonsense outputs -- are hypothesized to be caused by the \emph{exposure bias} \citep{bengio,ranzato}, which is understood as the inference-time inability to recover from own errors and that  manifests itself in a spectrum of erroneous translations from fluent, but semantically unrelated to the input, to completely non-linguistic outputs \citep{sennrich}. One suspected reason is that not all inference-time token decisions are featured in the training data and so there was no training signal to help to recover or to continue from them. This in turn, is believed to be caused by the standard teacher-forcing training algorithm \citep{oldrnnpaper}, which decomposes sequence learning into independent per-token predictions, each conditioned on the \emph{golden truth} context rather thauan the context the model would have produced. 

Several attempts tackle exposure bias by bringing the training and testing objectives closer, most notably with reinforcement learning (RL) \citep{ranzato, shen}. While RL does improve quality \cite{shen, kreutzer17}, to what extent this can be attributed to exposure bias reduction is not clear, as
RL generally replaces the learning algorithm causing the exposure bias, rather than using its proven mitigation. Moreover, fully-fledged RL is slower than maximum likelihood training (MLE) with teacher-forcing~\cite{shen}, which is a limiting factor in applications.

In this work\footnote{See extended report in~\cite{hormann21master}.}, we attempted to address exposure bias directly via a connection to robotics, where the same underlying problem of catastrophic control failures of \emph{behavioral cloning} (the equivalent of teacher-forcing) is well understood: behavioral cloning has been theoretically proven to amplify errors \citep{kaarinen, ross1} that explains its failures, and several imitation learning (IL) based mitigation strategies with guarantees exist. Specifically, we are motivated by the result of \cite{ross1}, that a system trained with behavioral cloning accumulates errors quadratically with the target length $T$ in the worst case. We further adapt their algorithm
AggreVaTe \citep{ross2} to NMT, that is free from error accumulation. Its major requirement 
is a powerful oracle capable of correcting arbitrary inference errors on-demand. Once such oracle is available, on-the-fly augmenting training data with the oracle's corrections allows the student NMT to be exposed to (optimal) corrections of its own errors, enjoying only linear worst-case error accumulation. 

The bulk of our work is devoted to the construction and evaluation of the oracle. Given an input $x$ and the reference $r$, it should be able to continue any (potentially, erroneous) partial translation $y_{1:t}$ in a BLEU-optimal way:
\setlength{\abovedisplayskip}{3pt}
\setlength{\belowdisplayskip}{3pt}
\begin{equation}
y^\ast_{t+1:T} = \argmax_{T, y_{t+1:T} s.t. y_{1:T} \in L} \text{BLEU}(y_{1:t} + y_{t+1:T}, r),\label{eq:prefix_oracle}
\end{equation}
where $y_{1:T}=y_{1:t} + y_{t+1:T}$ is the concatenation of the prefix, $y_{1:t}$, and the continuation, $y_{t+1:T}$, and $L$ is the search space which, in our case, will be a statistical machine translation (SMT) lattice.

Token cross-alignments make~\eqref{eq:prefix_oracle} a hard problem and many possible hypotheses are unreachable in $L$. However, finding such oracle translations -- reachable hypotheses closest to the references -- has been studied for SMT, where with the help of linear BLEU approximations \citep{tromble} and shortest path algorithms excellent oracle translations, almost doubling BLEU scores vs.\ regular SMT translations, could be found via a finite-state transducer (FST) representation of search space $L$~\cite{sokolov13lattice}. The main idea behind our IL oracle was to repurpose these existing BLEU oracles to solve~\eqref{eq:prefix_oracle}, powered by the observation that there exists a gap of >25 BLEU points between NMT quality and the oracle quality (Figure~\ref{fig:evaluation googleTranslate roll-ins}, left).

We report a positive and a negative result: First, the overhead of on-the-fly querying the SMT error-correcting oracle is tolerable and permits efficient IL implementation of NMT training. 
\todo{add slowdown comparison}
Second, however: while SMT oracles do find high-BLEU oracle translations when no (or a short) prefix is enforced, even the least pruned lattices are still overly reduced by the SMT decoder to keep only very reasonable translations and exclude many student prefixes which are though required to be in-lattice for the oracle to correct them. Although we implemented a workaround to fix this prefix unreachability problem, it did not lead to corrections improving over the student's own translations.


\section{Imitation Learning Training of NMT}\label{sec:il}

We treat an auto-regressive NMT system as a time-dependent policy $\pi$ that specifies a distribution over target tokens conditioned on input $x$ and previously generated prefix $y_{<t}$, both of which will be omitted for clarity: $\pi_t(y) \equiv \pi(y|y_{<t}; x)$. This policy is implemented as the output of the last softmax layer and depends functionally on the rest of the network\footnote{We also reuse the notation to denote whole or partial generated sequences as $\pi$'s output, e.g. $y_{1:T}=\pi(x)$.}. When training with teacher-forcing, the following loss is minimized under the \emph{empirical} distribution of parallel sequences in training data $\mathcal{D}$: $\mathcal{L_{\text{MLE}}(\pi)} = \mathbb{E}_{\mathbf{y}|\mathbf{x}\sim\mathcal{D}}[\sum_{t=1}^T \ell(y_t, \pi_t)]$, where $\ell(\cdot)$ is usually the negative log-likelihood. However, to perform well at test time we are interested instead in $\mathcal{L(\pi)} = \mathbb{E}_{\mathbf{y}|\mathbf{x}\sim\pi}[\sum_{t=1}^T \ell(y_t, \pi_t)]$, i.e. we need to perform well under the \emph{learned} model distribution. As shown by \citet{ross1}, the discrepancy between $\mathcal{L}$ and $\mathcal{L_{\text{MLE}}}$ can accumulate quadratically with the sequence length $T$, which in practice could manifest itself as catastrophic translation failures.

To mitigate, \citet{ross2} proposed the AggreVaTe algorithm, which is free from such error accumulation and would be an attractive asset for NMT. It, however, relies on an oracle policy $\pi^\ast$ that, given the same input $x$ and the partially generated $\pi$'s prefix $y_{<t}$, can produce the full continuation in the task-loss optimal fashion. For NMT, that translates into solving the BLEU optimization task~\eqref{eq:prefix_oracle}, which is our main technical contribution.

AggreVaTe aims to make the student's action-value function $Q$ (here, unnormalized logits before softmax) accurately predict the expert continuation's BLEU score for any action, i.e. vocabulary token. Each training example $x$ is translated by $\pi$ to get translation $y$; then an exploration action, $a_t$, is sampled at a random position $t$. The prefix $y_{<t}$ is concatenated with $a_t$ and continued by the oracle to obtain the BLEU score of this correction for the particular action $a_t$ given context $y_{<t}$; this score is then used as the target value for $Q$s.

\begin{algorithm}[h] 
\caption{AggreVaTe for NMT} 
\label{alg:the_algorithm} 
    \SetInd{0.45em}{0.45em}
    \KwData{data $\mathcal{D}$, oracle $\pi^\ast$}
    Initialize $\pi_{0}$\;
    \For{$j = 1\dots J$}{
        $\Lambda=\emptyset$\;
        \For{$x$ \textbf{in} $\mathcal{D}$}{
            Predict $y_{1:T}=\pi_{j-1}(x)$\;
            Sample uniformly $t \in \{1,...,T\}$\;
            Select an exploratory action $a_t$\;
            Solve~\eqref{eq:prefix_oracle}: $y^\ast_{t+1:T}=\pi^\ast(y_{<t}+a_t;x)$\;
            Augment $\Lambda$ with a loss term $\ell_j=I\cdot\big(\sigma(Q(a_t;y_{<t})\!-\!\Delta\!\operatorname{BLEU}(a_t)\big)^2$\; 
        }
        Train policy $\pi_j$ on $\Lambda$ with Adam\;
    }
\end{algorithm}
In Algorithm~\ref{alg:the_algorithm}, the sigmoid $\sigma$ scales logits to \mbox{[0, 1]} to match the BLEU range (its slope is chosen such that min and max values of $Q$ roughly map to 0 and 1, resp.), $I$ is the indicator testing that the oracle's continuations are of higher BLEU than the student's, and the ``reward-to-go'' at time $t$, $\Delta\operatorname{BLEU}(a_t)$, is defined as the contribution of the oracle's suffix $y^\ast_{t+1:T}$ to the total sequence's score:
$\operatorname{BLEU}(y_{<t}+a_t+y^\ast_{t+1:T})-\operatorname{BLEU}(y_{<t}+a_t))$.

\paragraph{Exploratory actions.}
After beam search on $\pi_j$ produces $y_{1:T}$ and a random $t$ is sampled, we need to generate an exploration action $a_t$. 
We evaluated three ways of selecting~$a_t$:
\begin{enumerate}
\itemsep0em 
    \item A uniform random action as in~\citep{ross2}. This leads to slower training as most tokens will not influence BLEU, but can eventually help generalization.
    \item The most probable action as per student's $\pi$. This uses the oracle more efficiently, focusing on correcting student errors that are about to be committed.
    \item A stochastic mixture of the above two methods with probability $\beta$ of selecting 1) or 2) otherwise, to balance their trade-offs.
\end{enumerate}
Empirically, mixing with $\beta=0.1$ worked best.
\section{Building the BLEU oracle}\label{sec:oracle}
To build an oracle solving~\eqref{eq:prefix_oracle}, 
we follow~\citet{sokolov13lattice}, who convert phrase lattices generated by Moses decoder~\cite{koehn} to FSTs and score edges with the linear BLEU approximation from \citep{tromble}: For a hypothesis $y$ and a reference $r$ the log-BLEU is assumed to be approximately linear in $n$-gram precisions:
\begin{equation}
\label{eq:linBLEU}
    \!\!\operatorname{BLEU}'(y,r)\!=\!\theta_{0}\left|y\right|\!+\!\sum_{n=1}^{N} \theta_{n}\!\!\sum_{u \in \Sigma^{n}}\!\! c_{u}\left(y\right) \delta_{u}(r),
\end{equation}
where $c_u(e)$ is the number of times the $n$-gram $u$ appears in $y$, $\delta_u(r)$ is an indicator for presence of $u$ in $r$; $\theta_n$ are further parametrized as $\theta_{0}=1, \theta_{n}=-\left(4 p \cdot r^{n-1}\right)^{-1}$, where $p$ and $r$ are grid-sought on a dev set to maximize corpus BLEU (\S\ref{sec:oracle_tuning}).

Given pre-decoded, reweighted with~\eqref{eq:linBLEU} and cached FSTs for every training example, shortest paths on them correspond then to the hypotheses with (approximately) the best BLEU in the FST w.r.t.\ to the corresponding reference, thus solving~\eqref{eq:prefix_oracle} \emph{from scratch} for $t=0$. 
For the IL use-case, however, the oracle must \emph{continue} hypotheses, rather than create them from scratch, so it must accept an existing prefix $y_{<t}$ and solve~\eqref{eq:prefix_oracle} with this constrain. Since $y_{<t}$ is not guaranteed to be reachable in $L$, we perform a two-step procedure: First, make all FST states final to allow termination of shortest paths in any state and set $r=y_{<t}$; then, the shortest path is a reachable partial translation, $y'_{<t}$, closest to $y_{<t}$ as per~\eqref{eq:linBLEU}. Second, prune all edges not reachable from $y'_{<t}$ from the original FST and \mbox{re-solve}~\eqref{eq:prefix_oracle}, this time, with the actual reference~$r$. 

Since~\citet{sokolov12computing} showed that by approximating bi-gram BLEU in~\eqref{eq:linBLEU} only a marginal decrease in corpus BLEU can be traded for an order of magnitude faster computation, we implemented the oracle for $N=2$ in C++ with Python bindings to enable calls from the \texttt{fairseq} NMT toolkit.

\paragraph{Oracle efficiency.} 
\begin{figure}
    \centering
    \includegraphics[width=0.75\columnwidth]{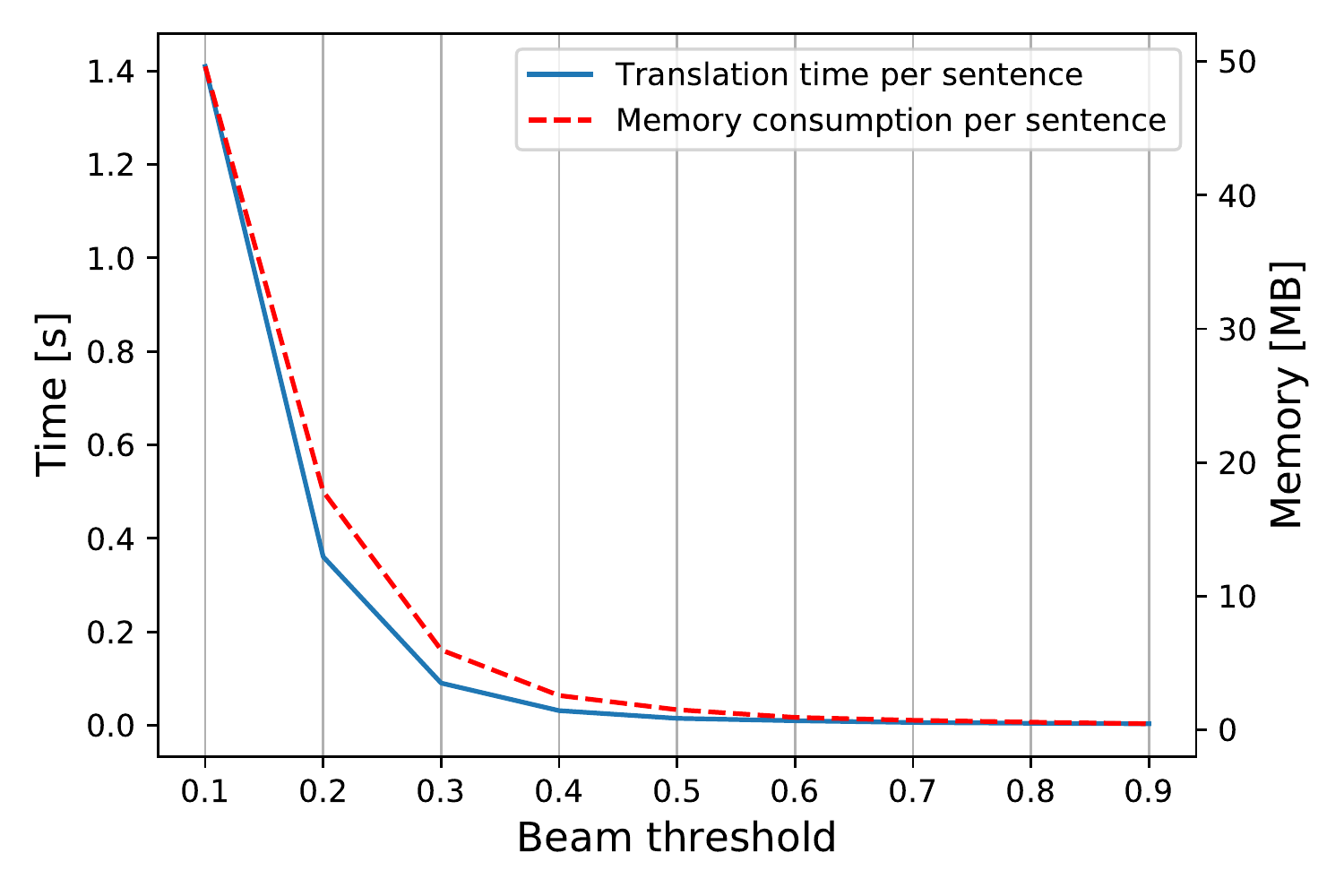}
    \caption{Oracle's average memory consumption and continuation time.}
    \label{fig:oracle_translation_performance}
\end{figure}
Figure~\ref{fig:oracle_translation_performance} shows that oracle's memory consumption decreases from 50Mb for Moses' lattice beam threshold $b=0.1$ (low pruning) to a 1/100th of that (10Kb) for $b=0.9$ (high pruning). Increasing $b$ also reduces the oracle continuation time as FSTs have fewer states and transitions: The shortest path takes from about 1 to 0.1 secs, allowing oracle queries on every SGD update. As oracle calls are self-contained and therefore highly parallelizable, time should be further divided by available CPUs.

\begin{figure}
    \centering
    \includegraphics[width=1.0\columnwidth]{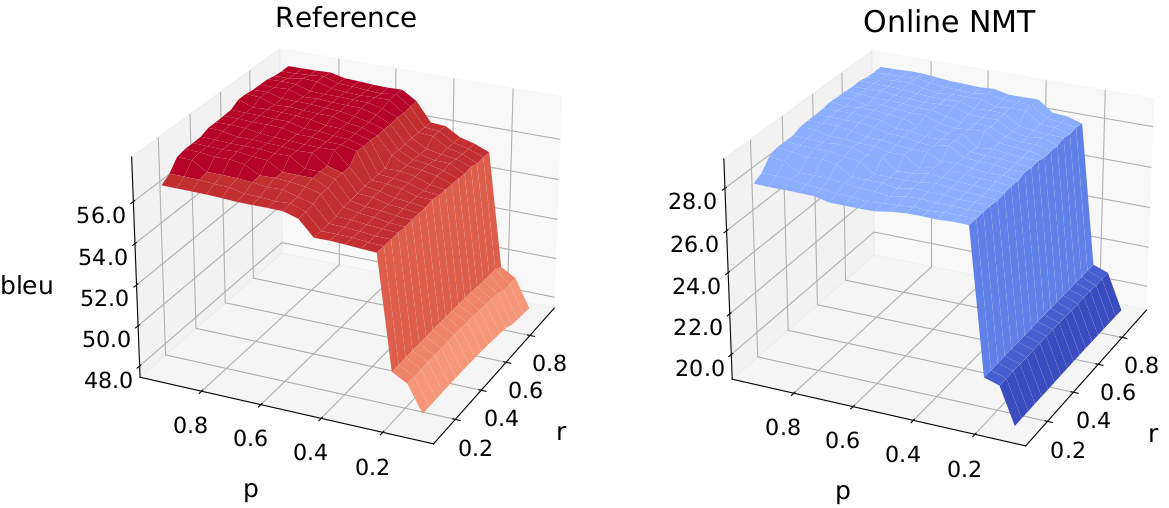}
    \caption{Oracle's BLEU for reference (right) and imperfect (left) prefixes for pruning threshold $b=0.1$ (other $b \in \{0.2,0.3,0.4\}$ give similar results). Baseline NMT achieves BLEU 32 on the same dev set.}
    \label{fig:evaluation googleTranslate roll-ins}
\end{figure}

\paragraph{Oracle quality.} 
Figure \ref{fig:evaluation googleTranslate roll-ins} (left) illustrates grid-search results for $p$ and $r$ with reference prefixes. As can be seen, the achievable oracle BLEU on the IWSLT14 EN-DE dev set exceeds the teacher-forcing NMT quality ($\sim$32 BLEU) by >25 points. 


\section{Experiments}\label{sec:training}
We experimented with the IWSLT14 DE-EN dataset tokenized, lowercased and cleaned with Moses tools, and jointly BPE-split into 32,000 tokens. Training hyperparameters are given in Table~\ref{tab:fundamental_hyperparameters} in Appendix. 
We warm-started fine-tuning with AggreVaTe from the baseline's last checkpoint; the learning rate was reduced to $5 \times 10^{-7}$ and its schedule changed to ``fixed''.

\begin{figure}
    \centering
    \includegraphics[width=0.9\columnwidth]{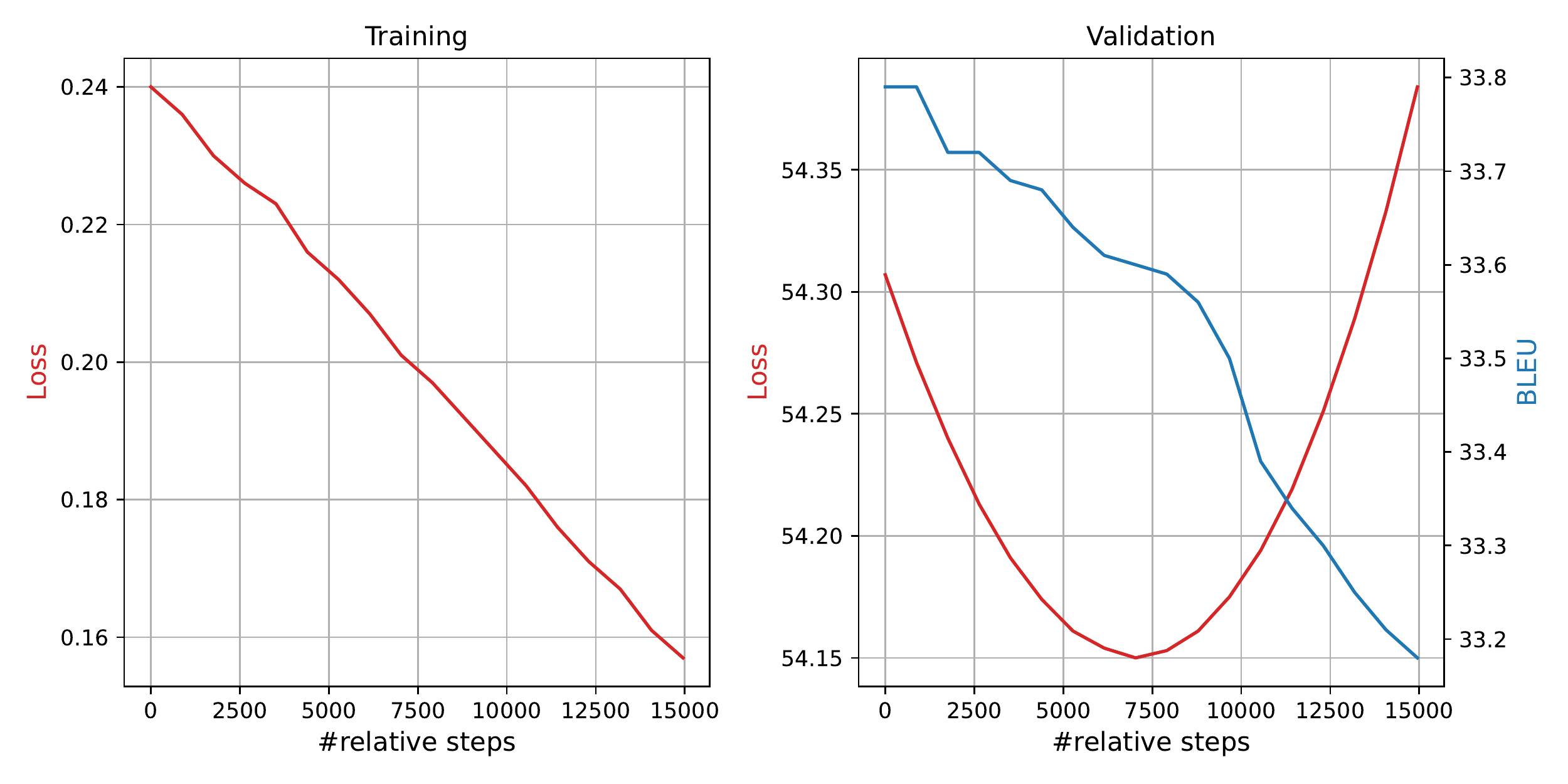}
    \caption{Training and validation loss/BLEU during $15k$ fine-tuning steps of a converged baseline. }
    \label{fig:objectives_training_validation}
\end{figure}

Despite the high performance of the oracle and working training loop we failed to improve over the teacher-forcing training. While in Figure~\ref{fig:objectives_training_validation} the training and validation losses decrease when warm-starting AggreVaTe, this is not mirrored by dev BLEU increase. Moreover, the same behaviour happens even for AggreVaTe on top of an unconverged student (Figure~\ref{fig:BLEU_irl_objectives} in \S\ref{sec:more_results}), hinting at systematic low-quality oracle outputs. More evaluations, including other quality metrics, can be found in \S\ref{sec:more_results}. 

Below we hypothesize and verify three related reasons that could explain this oracle failure to correct NMT student's errors despite high BLEU on the unconstrained oracle task. 

\subsection{Failure analysis}

\paragraph{Oracle overfits to references.}
The oracle achieves high BLEU for oracle translations when finding them from scratch, or when continuing prefixes, that are extracted from references, which are not representative of real student prefixes.
To confirm the hypothesis that the SMT oracle search spaces are overly bound to high-scoring translation, and to understand how prefix quality influences the oracle quality, we used an online NMT service to generate good but imperfect translations and derive our prefixes from them. 

While the optimal approximation hyperparameters, $p$ and $r$, did not visibly change when continuing imperfect prefixes, the drop in oracle performance vs.~reference prefixes are considerable 20+ BLEU points (Figure~\ref{fig:evaluation googleTranslate roll-ins}, right), confirming that oracle on suboptimal prefixes is underperforming. 

\paragraph{Non-monotonic search spaces.}

\begin{figure}
    \centering
    \includegraphics[width=0.95\columnwidth]{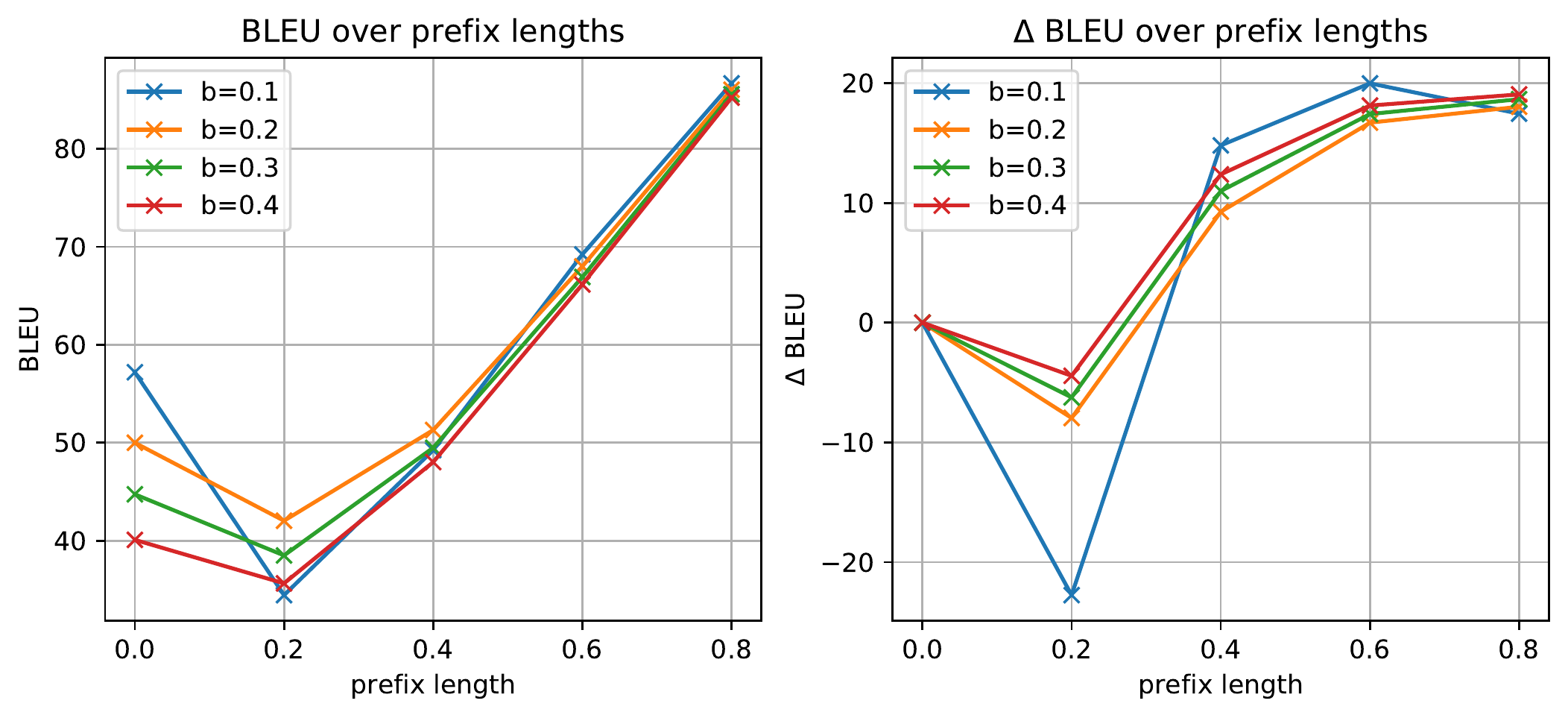}
    \caption[Prefix length influence on BLEU]{Left: BLEU score over prefix lengths taken from the reference. Right: Difference of BLEU score from one discrete prefix length to the next. }
    \label{fig:comparison gridsearch bleu over prefix lengths}
\end{figure}
Figure \ref{fig:comparison gridsearch bleu over prefix lengths} shows the influence of the prefix length and the beam threshold $b$ on corrections' BLEU. 
As can be seen, less pruning of the lattice (decreasing $b$) does not lead to improved oracle quality, meaning that, 
counter-intuitively, not every path that exists in a more pruned lattice (larger $b$) is also present in it for lower $b$. We hypothesize that the increased flexibility of larger lattices becomes a problem when the beam search, that drives the pruning, has more options to minimize the \emph{model} score on idiosyncratic paths that, due to model and search errors, do not necessarily maximize BLEU w.r.t.~the reference. This corroborates the findings of~\citet{sokolov13lattice} (Table~VIII), who found that smaller $b$ do not lead to a significant oracle BLEU increase despite an explosion of the number of edges. 

Unfortunately, this lattice non-monotonicity may disproportionally affect our oracle because of the two-step unreachability mitigation procedure (\S\ref{sec:oracle}). 

\paragraph{Lattices lack diversity.}
Our final hypothesis is that lattices do not match the diversity of NMT outputs; and so they struggle to correct an underperformant student, finding reachable prefixes $y'_{<t}$ that are wildly different from the actual $y_{<t}$. 
\begin{table}
    \centering\resizebox{\columnwidth}{!}{%
    \begin{tabular}{ll|cccc}
        \toprule
         $b$ & $\beta$    &  s\_BLEU &  s\_GLEU &  BLEU &  GLEU \\
\midrule
        0.1 & 0.1 &        13.59 &        11.56 & 29.16 & 31.58 \\
            & 0.5 &        12.86 &        10.97 & 27.19 & 29.74 \\
            \midrule
        0.2 & 0.1 &        13.01 &        11.17 & 28.23 & 30.87 \\
            & 0.5 &        11.89 &        10.26 & 26.12 & 28.95 \\
            \midrule
        0.4 & 0.1 &        10.76 &         9.38 & 25.95 & 29.07 \\
            & 0.5 &         9.88 &         8.70 & 23.88 & 27.20 \\
        \bottomrule
    \end{tabular}}
    \caption[Effects of beam threshold and exploration randomness on scores]{BLEU/GLEU  grouped by beam threshold $b$ and exploration randomness $\beta$.}
    \label{tab:Student_beam-threshold_exlo-rand}
\end{table}
In Table~\ref{tab:Student_beam-threshold_exlo-rand} we measure BLEU and GLEU as a function of $b$ and increased exploration $\beta$. As the student should generate better-than-random tokens, replacing them more often with random ones should significantly decrease oracle continuation scores. We observe, however, a relatively modest (for an oracle) drop, pointing to difficulties of continuing reasonable student's prefixes even without random exploration. To make sure that the drop is not due to the inserted random token only and affects the whole continuation, we also report the suffix-BLEU/GLEU (denoted with `s\_') that measures the suffix contribution to the total metric (see $\Delta\operatorname{BLEU}(a_t)$ in Algorithm~\ref{alg:the_algorithm}). More results in~\S\ref{sec:diversity}.




\bibliography{anthology,custom}
\bibliographystyle{acl_natbib}

\clearpage
\appendix

\section{Oracle tuning}\label{sec:oracle_tuning}
Grid-searches for $p$ and $r$ were done in \mbox{[0.1, 0.95]} with the step 0.05, the Moses beam pruning threshold $b$ -- in \mbox{[0.1, 0.9]} with the step 0.1. Prefixes $y_{<t}$ were randomly cut from references or student translations at length \mbox{[0\%, 80\%]} with the 20\% step.

\section{More Evaluation Results}
\label{sec:more_results}

\begin{figure}
    \centering
    \includegraphics[width=1.0\columnwidth]{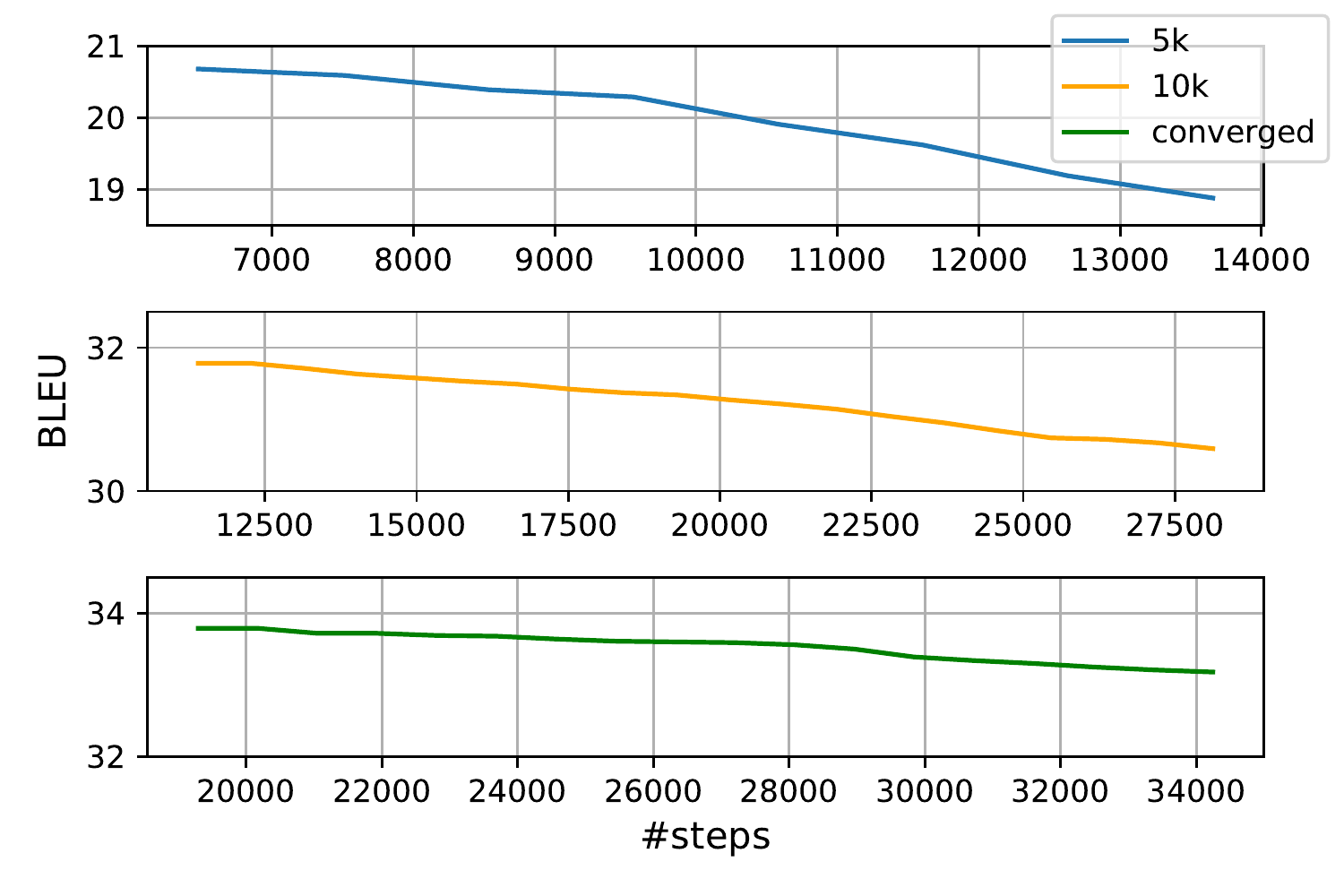}
    \caption[BLEU performance for different starting points of fine-tuning]{BLEU performance on validation set during fine-tuning with AggreVaTe at different starting points.}
    \label{fig:BLEU_irl_objectives}
\end{figure}
\todo{Fig. \ref{fig:BLEU_irl_objectives} remove BSE curve, rename S-BSE to AggreVaTe or w/o name}

\paragraph{Fine-tuning at different stages of convergence.} Figure \ref{fig:BLEU_irl_objectives} shows the effect of fine-tuning the baseline model from the starting points 5k, 10k and 17.5k steps respectively. The experiments were stopped after 8 hours of fine-tuning. Training is decreasing the BLEU score on the validation set with increasing number of training iterations.

\paragraph{Other metrics.} In Table \ref{tab:comparison baseline s-bse finetune} shows the evaluation on the IWSLT14 test set for four additional metrics, known to better correlate with human scores.
\begin{table}
    \centering
    \resizebox{\columnwidth}{!}{
    \begin{tabular}{l|c|c|c}
    \toprule
        Metric & Baseline& Fine-tuned best & Fine-tuned last \\
         & 17.5k & 26.3k & 34.2k \\
        \midrule
        BLEU & \textbf{34.34} & 34.06 & 33.66 \\
        GLEU & \textbf{0.665} & 0.663 & 0.66 \\
        METEOR & \textbf{0.57} & 0.569 & 0.564 \\
        BLEURT & 0.17 & \textbf{0.173} & 0.172 \\
        Perplexity & 6.11 & \textbf{6.06} & 6.11 \\
        \bottomrule
    \end{tabular}
    }
    \caption[Baseline compared against fine-tuned model in MT metrics]{Comparison of the baseline and fine-tuned AggreVaTe models on test set. The best model was selected according to the NLL loss on the validation set.}
    \label{tab:comparison baseline s-bse finetune}
\end{table}
The \textit{Fine-tuned best} was trained for $26.3k - 17.5k = 8.8k$ steps and selected by the minimum NLL loss on the validation set from the models during fine-tuning. The \textit{Fine-tuned last} model is fine-tuned for additional $34.2k - 17.5k = 16.7k$ steps.

In the experiments, the best fine-tuned model exceeds the performance of the last fine-tuned model in all of the measured metrics. Compared to the baseline, the best fine-tuned model performs slightly better in BLEURT and perplexity but fails to improve BLEU, GLEU and METEOR.

\paragraph{Measuring effect on exposure bias.}
\begin{figure}
    \centering
    \includegraphics[width=0.9\columnwidth]{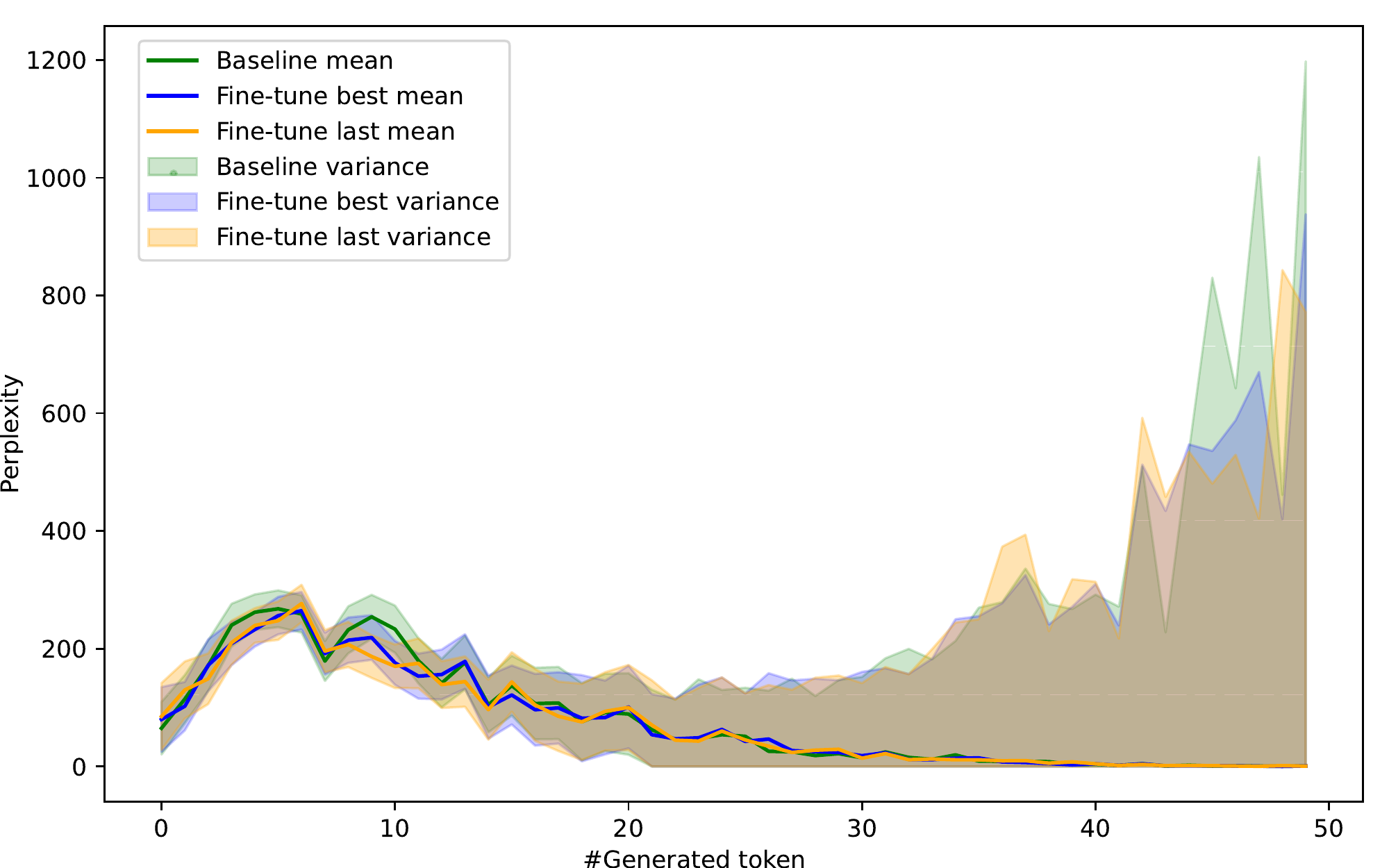}
    \caption[Perplexity over increasing number of generated tokens]{Mean and variance of perplexity over number of generated tokens for baseline and fine-tuned models.}
    \label{fig:perplexity_over_token}
\end{figure}
In Figure~\ref{fig:perplexity_over_token} we measured the mean and variance of the perplexity for each consecutive token for the test setting. All models are pretty certain in the beginning of the sequence and get more uncertain after around 10 tokens, which quantifies the exposure bias phenomenon.
For the baseline model, this is expected, among other things, because for initial tokens the self-attention has less tokens to attend to; the increased variance after 30 generated tokens can be explained by inability to correct own errors, as well as the low number of training examples actually reaching that length. As \mbox{AggreVaTe} training failed, we do not observe improvements in perplexity for fine-tuned models.

\section{Lattice diversity}\label{sec:diversity}
\begin{figure}
    \centering
    \includegraphics[width=0.9\columnwidth]{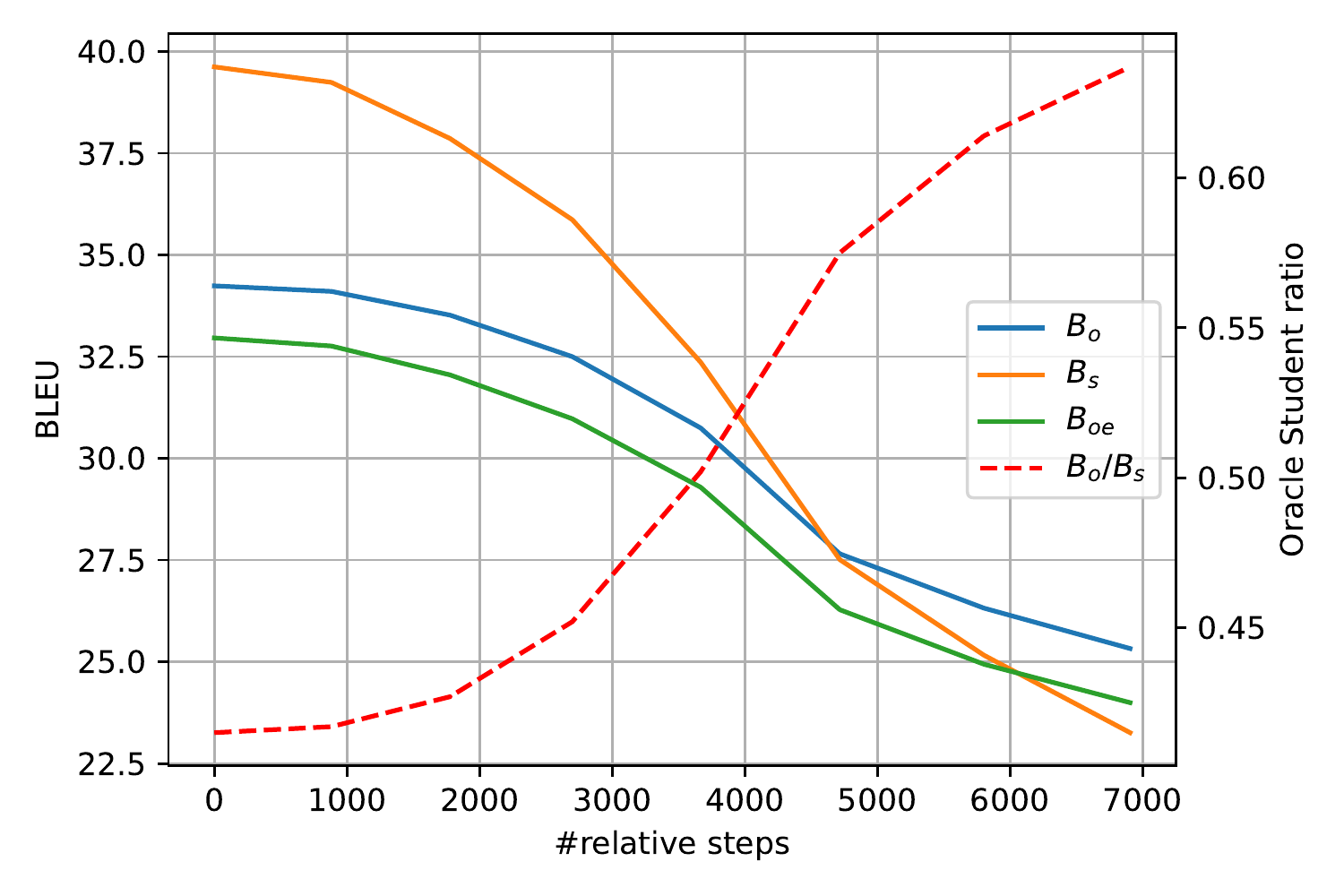}
    \caption{BLEU performance of student ($B_s$), oracle ($B_o$), oracle with student exploration action ($B_{oe}$) and the ratio of oracle to student BLEU ($B_o/B_s$) over the course of fine-tuning.}
    \label{fig:BLEU_sub_scores_over_training}
\end{figure}

To investigate the lattice diversity further, we plotted in~Figure~\ref{fig:BLEU_sub_scores_over_training}
the following averages: the BLEU score of the oracle ($B_o$), student ($B_s$), of the oracle's using the student's exploration action ($B_{oe}$) and the ratio $B_o/B_s$ to show for how many sentences the oracle's continuations were better than the student's inferred sentences (corresponds to the indicator $I$ in Algorithm~\ref{alg:the_algorithm}).

As the student's performance gets worse due to training on sub-optimal examples, $B_o/B_s$ increases which is a sign of the student dropping in quality below oracle corrections. However, the $B_{oe}$ curve, which includes the exploration action, is below $B_s$ most of the time, witnessing that the oracle is confused both by the exploration action as well as by the student's prefix (after $B_s$ decreases below $B_{o}$ or $B_{oe}$, $B_s$ still does not acquire positive slope).

\begin{table}
    \centering
    \resizebox{\columnwidth}{!}{%
    \begin{tabular}{l|c|c}
    \toprule
        Parameter & Baseline & Fine-tuning \\
        \midrule
        Learn. Method & Teacher-Forcing & AggreVaTe \\
        Criterion & Cross-Entropy & MSE \\
        Optimizer & Adam (default) & \# \\
        Learn. Rate & $5\times 10^{-4}$ & $5\times 10^{-7}$\\
        Learn. Rate Scheduler & Inv. Square Root & Fixed \\
        Dropout & $0.3$ & \# \\
        Weight Decay & $1\times 10^{-4}$ & \# \\
        Warmup  & $4000$ & $0$ \\
        Tokens per Batch & $4096$ & \#\\
        \bottomrule
    \end{tabular}
    }
    \caption[Baseline and fine-tune hyperparameters]{Hyperparameters for training the MLE baseline and fine-tuning with AggreVaTe. Unchanged parameters are marked with \#.
}
    \label{tab:fundamental_hyperparameters}
\end{table}

\end{document}